\documentclass{article}

\usepackage{microtype}
\usepackage[final]{colm2026_conference}
\usepackage{graphicx}
\usepackage{subcaption}
\usepackage{colortbl}
\usepackage{booktabs} %
\usepackage{float}

\usepackage{hyperref}
\usepackage{url}
\usepackage{lineno}

\usepackage{amsmath}
\usepackage{amssymb}
\usepackage{mathtools}
\usepackage{amsthm}
\usepackage{xspace}
\usepackage[capitalize,noabbrev]{cleveref}
\theoremstyle{plain}

\theoremstyle{definition}

\theoremstyle{remark}

\newcommand{\ours}{{\sc NeSyS}\xspace}
\usepackage[inline]{enumitem}
\usepackage{tcolorbox}
\tcbuselibrary{breakable}

\definecolor{darkblue}{rgb}{0, 0, 0.5}
\hypersetup{colorlinks=true, citecolor=darkblue, linkcolor=darkblue, urlcolor=darkblue}

\title{Neuro-Symbolic Synergy for World Modeling}

\author{
Hongyu Zhao \\
Department of Computer Science \\
University of Maryland \\
College Park, MD 20742, USA \\
\texttt{hongyuz@umd.edu} \\
\And
Siyu Zhou \\
Faculty of Engineering and IT \\
Australian AI Institute \\
University of Technology Sydney \\
Sydney, Australia \\
\And
Haolin Yang \\
Department of Statistics \\
University of Chicago \\
Chicago, IL 60637, USA \\
\And
Zengyi Qin \\
Department of Computer Science \\
Massachusetts Institute of Technology \\
Cambridge, MA 02139, USA \\
\AND
Tianyi Zhou \\
Mohamed bin Zayed University of Artificial Intelligence \\
Abu Dhabi, UAE \\
\texttt{tianyi.david.zhou@gmail.com} \\
}

\begin{document}

\ifcolmsubmission
\linenumbers
\fi

\maketitle

\begin{abstract}
Large language models (LLMs) exhibit strong general-purpose reasoning capabilities, yet they frequently hallucinate when used as world models (WMs), where strict compliance with deterministic transition rules—particularly in corner cases—is essential. In contrast, Symbolic WMs provide logical consistency but lack semantic expressivity. To bridge this gap, we propose \emph{Neuro-Symbolic Synergy} (\ours), a framework that integrates the probabilistic semantic priors of LLMs with executable symbolic rules to achieve both expressivity and robustness. \ours alternates training between the two models using trajectories inadequately explained by the other.
Unlike rule-based prompting, the symbolic WM contributes candidate-level scores through log-linear reranking, without requiring the LLM to interpret rule text. Rule-guided sampling prioritizes transitions that are weakly covered by symbolic rules, using 35--60\% of the training pairs while outperforming full-data supervised fine-tuning in five of six settings.
Experiments on ScienceWorld, WebShop, and PlanCraft demonstrate consistent gains in WM prediction accuracy and data efficiency; one-step lookahead on open-ended WebShop also improves agent reward.
Our models, rules, and code are available at \url{https://github.com/tianyi-lab/NeSyS}.
\end{abstract}

\begin{figure}[ht]
    \centering
    \includegraphics[width=0.7\textwidth]{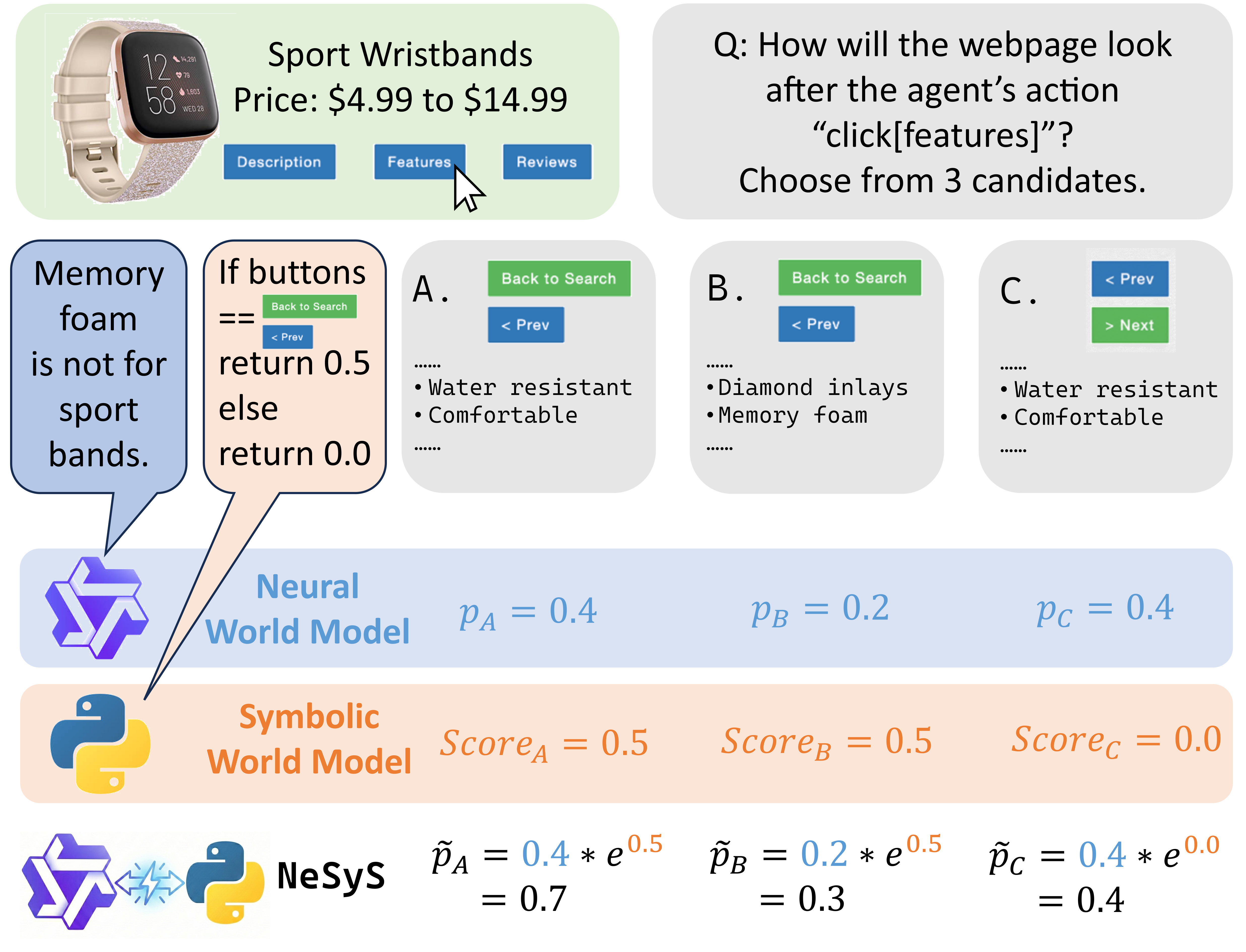}
    \caption{An example of the world modeling task. Given the current belief state and the agent's next action, the world model needs to predict the next state. Both a neural world model (LLM) and a symbolic world model fail to answer the question by themselves. We combine their scores through candidate-level log-linear reranking.}
    \label{fig:cover}
\end{figure}

\section{Introduction} \label{sec:intro}
Large language models (LLMs) have emerged as general-purpose world models (WMs) for sequential decision-making~\citep{yang2024evaluating}, yet they often struggle in domains with well-defined dynamics such as games and web environments~\citep{levy2025worldllm, chae2024web}. In particular, they frequently fail in deterministic scenarios requiring strict adherence to transition rules~\citep{duan2024gtbench}. Although extensive domain-specific fine-tuning can partially mitigate these issues, the inherently probabilistic nature of LLMs makes hard-constraint compliance difficult to guarantee, and their statistical learning limits the ability to capture rare corner cases and long-tail behaviors.

In parallel, explicit symbolic WMs~\citep{agravante2023learning,liang2024visualpredicator} enforce hard rules and deterministic transitions, naturally avoiding constraint violations. However, they struggle to generalize to complex, high-dimensional, or stochastic environments. This limitation becomes particularly severe in natural-language-driven environments, where predicting the exact next state requires the rich semantic priors of LLMs.

Neither purely neural nor purely symbolic WMs suffice alone, motivating \textbf{neuro-symbolic synergy} (\ours) in world modeling. \Cref{fig:cover} illustrates this with an example in which \ours correctly answers a question that both approaches fail to resolve on their own. Prior neuro-symbolic methods such as WALL-E~\citep{zhou2024wall} inject symbolic rules into the LLM's input context, but this does not reliably transfer to smaller or fine-tuned models that frequently ignore prompt-injected instructions.

In contrast, we propose to \textbf{directly rerank candidate probabilities with scores from symbolic rules implemented by Python functions}. This eliminates any reliance on instruction-following quality. Moreover, rules can be evaluated efficiently and locally---without additional LLM inference---making rule application substantially more efficient than context injection. The log-linear combiner itself is standard; our contribution is the framework for learning complementary symbolic world-model rules and integrating them at the score level.

Furthermore, we introduce \textbf{rule-guided data selection}, which down-samples training examples already captured by learned rules, substantially reducing neural training data while maintaining strong performance. Experiments on ScienceWorld~\citep{scienceworld2022}, WebShop~\citep{yao2022webshop}, and PlanCraft~\citep{dagan2024Plancraft} show gains across diverse environments and backbone LLMs.

In summary, this work makes the following contributions:
\begin{itemize}
    \item We introduce \emph{Neuro-Symbolic Synergy} (\ours), a framework that integrates LLM-based world models with executable rules through score-level reranking, allowing rules to impose strong penalties without relying on instruction following.
    \item We introduce a complementary training paradigm in which neural training prioritizes weakly covered transitions and rule induction targets neural-model errors, substantially reducing finetuning data requirements without sacrificing performance.
    \item We empirically demonstrate that \ours achieves consistent performance improvements across diverse interactive environments and model scales.
\end{itemize}

\section{Related works}
\paragraph{Neuro-symbolic world models.} %
To mitigate inherent limitations of pure neural models---such as hallucination---neuro-symbolic world models combine neural representations with explicit structure~\citep{cano2025neurosymbolic,sehgal2023neurosymbolic,balloch2023neuro}. Existing methods use semantic parsers~\citep{agravante2023learning} or VLMs~\citep{liang2024visualpredicator} to obtain symbolic dynamics. CLIN stores induced causal abstractions in textual memory~\citep{majumder2023clin}, while WALL-E injects learned natural-language rules into the LLM context~\citep{zhou2024wall}. WALL-E~2.0 compiles knowledge into executable functions, but returns their outputs to the LLM as contextual feedback~\citep{zhou2025wall}; the model must still interpret that feedback. In contrast, \ours applies executable outputs directly to candidate scores. PSALM instead infers PDDL-style action semantics for a symbolic planner and falls back to an LLM when planning fails~\citep{zhu2024psalm}; \ours targets text-rich environments whose semantics cannot be exhaustively represented by a compact action theory.

\paragraph{Training data selection.} Data selection is critical for optimizing the performance of fine-tuned LLMs while minimizing computational costs~\citep{zhou2023lima}. While recent literature has explored rule-based methods to identify high-quality training subsets~\citep{li2024rule}, our approach diverges in its objective. Instead of filtering for general data quality, we aim to maximize the complementary strengths of the neural model and the symbolic components by selecting instances that are difficult for symbolic rules to solve.

\paragraph{Constrained decoding and reranking.} Constrained decoding modifies token-level distributions during generation, commonly to enforce syntactic formats~\citep{willard2023efficient,beurer2024guiding}, and has been extended to semantic and logical constraints~\citep{nagy2026chopchop,ma2025logically}. \ours instead modifies the final distribution over candidate next states. This is less expressive than token-level control, but is easier to learn and apply in our setting.

\section{Problem setting}\label{sec:problem}
We study sequential decision-making problems in a partially observable Markov decision process (POMDP). At step $t$, the agent needs to do action $a_t$ based on the current belief state $b_t$. It will then receive the next state $s_{t+1}$ and a reward $r_t$ from the environment.  \footnote{We do not distinguish observation and state here. See \cref{app:pomdp} for a rigorous definition.}

A \textbf{world model} is defined as a predictive model that estimates the next state $s_{t+1}$ and reward $r_t$ conditioned on the current belief state $b_t$ and action $a_t$. Following prior LLM-based agent work, we approximate the belief state by a textual context constructed from the task description $g$ and a truncated history of recent observations, actions, and rewards that fit within the model’s context window.

\section{Method}\label{sec:method}
Our framework consists of two world models (WM): Neural WM and Symbolic WM. Neural WM is in the form of an LLM, while Symbolic WM is a weighted set of Python rules. %

\subsection{Framework overview}
\begin{figure}[ht]
        \centering
        \includegraphics[width=0.45\textwidth]{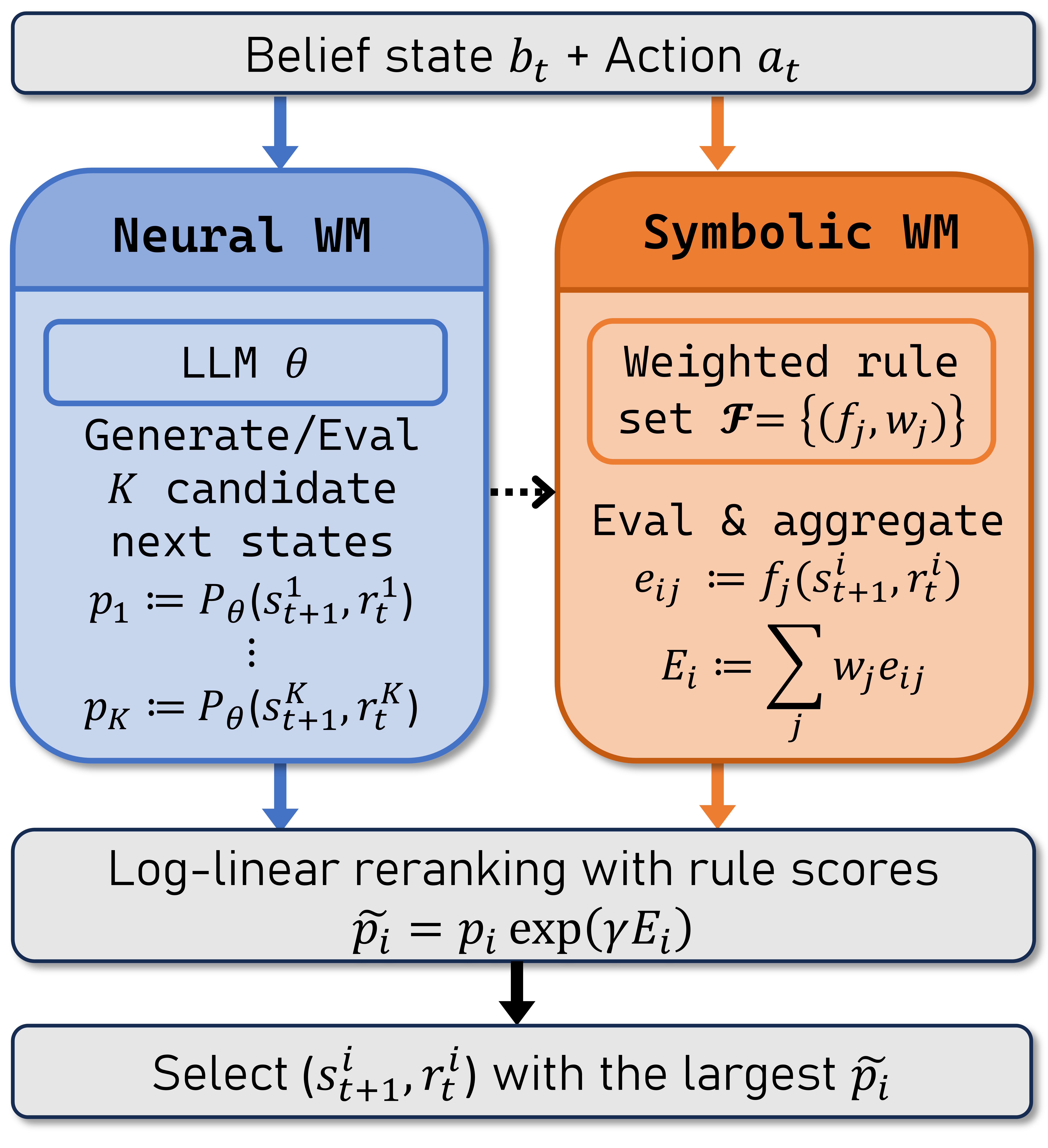}
        \caption{\textbf{Overview of \ours}. It consists of two world models: Neural WM and Symbolic WM. They are implemented as an LLM $\theta$ with likelihood function $P_\theta$ and a weighted rule set $\mathcal{F}$, respectively. Neural WM generates $K$ candidate next-state and reward pairs when choices are not provided and computes each likelihood $p_i$. Symbolic WM aggregates the score $e_{ij}$ produced by each rule $f_j$. We modify $p_i$ with the symbolic score and choose the candidate with the largest modified likelihood $\tilde{p}_i$.}
        \label{fig:inference}
\end{figure}

As mentioned in \cref{sec:intro}, the main idea of our strategy is to directly modify the probability distribution of Neural WM, i.e., LLM, with the symbolic rules learned by Symbolic WM. Given current belief state $b_t$ and action $a_t$, we aim to select the optimal next-step state and reward. The whole inference pipeline is illustrated in \cref{fig:inference}.

On the left side, Neural WM computes the likelihood ${p_i}$ for candidate next state $s_{t+1}^{i}$ and reward $r_t^{i}$. The $K$ candidates are either provided or generated by Neural WM itself. 

On the right side, Symbolic WM is implemented as a weighted set of executable Python functions $\mathcal{F}=\{(f_j, w_j)\}_{j=1}^m$, where $f_j\!:\!(b_t,a_t,s_{t+1},r_t)\mapsto\mathbb{R}$. Each rule produces a real-valued score $e_{ij}$ indicating how likely the candidate $(s_{t+1}^i,r_t^i)$ is correct under this specific rule. Rule synthesis requests scores in $[-1,1]$, while inference consumes the executable output as returned. Rules are typically inactive and output zero, but become active only when specific conditions are met (e.g., a particular action is taken or a relevant keyword appears in the belief state). We aggregate these logical judgments into a scalar shifting factor $E_i$ using learned rule weights $w_j$: $$E_i = \sum_{j=1}^m w_j e_{ij}$$

We combine these signals through log-linear reranking. The unnormalized modified score $\tilde{p}_i$ is
\begin{equation*}
\tilde{p}_i = p_i\exp(\gamma E_i),
\end{equation*}
where the default scale adapts to the candidate set, $\gamma=1+\max_k\log p_k-\min_k\log p_k$.

We select the candidate with the largest $\tilde{p}_i$. A negative $E_i$ imposes a strong but soft rule penalty: it does not mathematically zero the candidate's probability. Alternative combiners and the choice of $\gamma$ are examined in \cref{app:new-results}.

\subsection{Training pipeline}\label{sec:training}
\begin{figure}[ht]
        \centering
        \includegraphics[width=0.72\textwidth]{images/training_pipeline_camera_ready.png}
        \caption{\textbf{Training pipeline of \ours}. In Phase~1, development-set errors from the pretrained Neural WM drive rule induction; rules are accepted and calibrated only when they improve development performance. In Phase~2, we retain every training pair with zero active rules and sample covered pairs with probability proportional to $1/k$, where $k$ is the number of active rules on the gold candidate. The sampled data fine-tune Neural WM, and its residual development errors drive rule refinement.}
        \label{fig:training}
\end{figure}
As illustrated in \cref{fig:training}, our training pipeline consists of two phases: Initialization and Reciprocal Refinement. It uses development-set errors for rule induction, down-samples well-covered training transitions, and improves the two WMs based on each other's residual errors. This design encourages the modules to evolve complementarily rather than redundantly.

\subsubsection{Phase 1: initialization.}
\paragraph{Neural WM.}
We initialize Neural WM using a pretrained LLM. By evaluating this model on the development set, we can separate questions that can be solved with general common sense knowledge (correct samples) from those that require task-specific knowledge (incorrect samples). 

\paragraph{Symbolic WM.}
We aim to capture task-specific deterministic dynamics that Neural WM misses via explicit rules. The rules are induced via an automated debugging loop:
\begin{enumerate}
\item \textbf{Error clustering:} We identify mistakes made by the initialized Neural WM on a development set. These are clustered by syntactic and semantic similarity (details in \cref{app:clustering}) to identify systematic failure modes.
\item \textbf{Rule generation:} For each cluster, we prompt \texttt{gpt-5-mini} to write a Python function that corrects the specific error while generalizing to similar cases.
\item \textbf{Verification \& weighting:} A rule is added to $\mathcal{F}$ only if it improves \ours's development accuracy, filtering harmful or incorrect rules. We perform up to 3 reflection iterations with \texttt{gpt-5-mini} for failed rules; new rules receive an initial scalar weight $w_j=1$ before phase-specific calibration.
\end{enumerate}

\subsubsection{Phase 2: reciprocal refinement.}
\paragraph{Neural WM.}
A key insight of our work is that fine-tuning a model on behaviors already captured by explicit rules is computationally wasteful and potentially redundant. To address this, we introduce \textbf{rule-guided data selection}. 

We evaluate the training trajectories using our initialized rule set $\mathcal{F}$. For each gold transition, $k$ is the number of rules in $\mathcal{F}$ that return a non-zero score on its correct next-state and reward pair. Steps with smaller $k$ are harder to cover with the Symbolic WM alone. We retain all steps with $k=0$ and sample from the remaining steps with probability proportional to $1/k$, so that harder steps are sampled more frequently. Neural WM is then fine-tuned only on the sampled examples. This focuses the Neural WM's capacity on capturing complex, intuitive dynamics that are difficult to formalize as code. In practice, we find this strategy can discard about half of the training data while maintaining or improving performance.

\paragraph{Symbolic WM.}
Following refinement of Neural WM, rules that were useful for a weaker Neural WM might now negatively impact the performance. We first apply a cleaning pass: any rule that has a negative impact on the performance of Neural WM is discarded. We then add more rules by repeating the clustering and synthesis process described in Phase 1 to cover any new residual errors.

At the end of each phase, we calibrate scalar weights $w_j$ by coordinate descent on the development set; the Phase~1 and Phase~2 rows use weights calibrated for their corresponding rule sets. Weights only calibrate rule contributions; their development-set gain over uniform weights averages 2.6 points (range 0.3--4.8). After selection and weighting, all rules and weights are frozen and evaluated once on the unseen test set.

\section{Experiments} \label{sec:exp}
\subsection{Experiment setup} \label{sec:exp_setup}
\paragraph{Environments and data.} We evaluate on ScienceWorld~\citep{scienceworld2022}, WebShop~\citep{yao2022webshop}, and PlanCraft~\citep{dagan2024Plancraft}, which respectively require physical reasoning, web interaction, and structured crafting. Their full training splits contain 14,789 / 13,488 / 6,083 $(s,a,s')$ pairs from 1,088 / 1,257 / 1,198 trajectories. We construct multiple-choice development and test questions as detailed in \cref{app:data-collection}.
\paragraph{Settings and baselines.} We use Llama-3.2-1B-Instruct~\citep{meta2024llama32} and Qwen3-4B~\citep{qwen3} as neural WMs. \texttt{GPT-5-mini}~\citep{singh2025openai} is called only for offline rule induction; it is absent from the deployed WM and never called at inference. We compare with larger LLMs, direct-use GPT-5-mini, full SFT, and WALL-E~\citep{zhou2024wall}; details and additional baselines are in \cref{app:new-results}. We report each WM and \ours in both phases, using the gap-based $\gamma$ above and the development-calibrated weights for the corresponding phase.

\begin{table}[ht]
  \centering
  \resizebox{\textwidth}{!}{%
  \begin{tabular}{lcccccccc}
    \toprule
    \cmidrule(r){1-2}
    World Model     & Matter & Chem. & Class. & Biology & Forces & Meas. & Elec. & Avg.\\
    \midrule
    \rowcolor{gray!15}
    \multicolumn{9}{c}{\bf Baseline Models} \\
    \midrule
    Llama3.1-8B & 36.1 & 29.4 & 68.8 & 42.7 & 38.9 & 52.5 & 33.9 & 44.4 \\
    Qwen3-8B & 38.9 & 31.4 & 65.0 & 43.3 & 66.7 & 45.8 & 40.7 & 46.5 \\
    Qwen3-14B & 45.8 & 43.1 & 68.8 & 46.5 & 64.9 & 47.5 & 49.2 & 51.2 \\
    Phi-4-mini & 38.9 & 39.2 & 63.8 & 50.3 & 50.0 & 40.7 & 33.9 & 46.7 \\
    GPT-oss-20B &34.7& 49.0& 48.8& 47.7& 40.6& 33.9& 25.4& 41.2\\
    GPT-5-mini & 45.8 & 51.0 & 72.5 & 52.2 & 72.2 & 50.8 & 50.8 & 55.4\\
    \midrule
    \rowcolor{gray!15}
    \multicolumn{9}{c}{\bf Llama3.2-1B} \\
    \midrule
    SFT (100\% data)& 37.5 & 64.7& 81.3& 68.8& 80.6 & \textbf{54.2} & \textbf{62.7} & 64.4\\
    \midrule
    Neural WM (Phase 1) & 27.8 & 17.7 & 42.5 & 39.5 & 44.4 & 27.1 & 15.3 & 32.3\\
    Symbolic WM (Phase 1) & 33.3 & 39.2 & 58.8 & 56.7 & 41.7 & 30.5 & 47.5 & 46.9\\
    WALL-E & 31.9 & 27.5 & 56.2 & 42.7 & 41.7 & 33.9 & 22.0 & 38.3 \\
    \ours\ (Phase 1) & 43.1 & 27.5 & 61.3 & 58.6 & 50.0 & 42.4 & 37.3 & 48.8\\
    \midrule
    Neural WM (Phase 2) & 37.5 & \textbf{66.7} & 71.3 & 70.1 & 77.8 & 47.5 & 55.9 & 61.7\\
    Symbolic WM (Phase 2) & 45.8 & 41.2 & 78.8 & 56.7 & 41.7 & 45.8 & 55.9 & 54.7\\
    WALL-E & 38.9 & 58.8 & 73.8 & 66.2 & 83.3 & 49.2 & 44.1 & 59.5 \\
    \ours~(Phase 2, 45\% data) & \textbf{47.2} & 64.7 & \textbf{88.8} & \textbf{73.3} & \textbf{91.7} & 50.9 & 59.3 & \textbf{68.3}\\
    \midrule
    \rowcolor{gray!15}
    \multicolumn{9}{c}{\bf Qwen3-4B} \\
    \midrule
    SFT (100\% data) & 48.6 & \textbf{66.7} & \textbf{91.3} & 68.8 & \textbf{100.0} & \textbf{59.3} & 50.8 & 68.3 \\
    \midrule
    Neural WM (Phase 1) & 38.9 & 35.3 & 63.8 & 43.3 & 63.9 & 42.4 & 47.5 & 46.9 \\
    Symbolic WM (Phase 1) & 38.9 & 47.1 & 52.5 & 61.2 & 41.7 & 33.9 & 54.2 & 50.0 \\
    WALL-E & 47.2 & 27.5 & 67.5 & 47.8 & 83.3 & 44.1 & 47.5 & 50.8 \\
    \ours\ (Phase 1) 
    & \textbf{52.8} & 43.1 & 68.8 & 62.4 & 63.9 & 55.9 & 54.2 & 58.6\\
    \midrule
    Neural WM (Phase 2) & 50.0 & 58.8 & \textbf{91.3} & 66.9 & 94.4 & \textbf{59.3} & \textbf{66.1} & 68.5\\
    Symbolic WM (Phase 2) & 43.1 & 58.8 & 68.8 & 62.4 & 75.0 & 44.1 & 55.9 & 58.4 \\
    WALL-E & 48.6 & 60.8 & 86.2 & 70.1 & 91.7 & 55.9 & \textbf{66.1} & 68.1 \\
    \ours~(Phase 2, 45\% data) & 50.0 & 60.8 & 88.8 & \textbf{75.8} & 94.4 & \textbf{59.3} & \textbf{66.1} & \textbf{71.0}\\
    \bottomrule
  \end{tabular}
  }
  \caption{World modeling accuracy on ScienceWorld. \textbf{Bold} numbers highlight the best performance for each backbone model; data percentages give the fraction of the full training split used for neural fine-tuning.}
  \label{tab:wm_science_world}
\end{table}

\subsection{ScienceWorld}\label{sec:scienceworld}
ScienceWorld requires agents to navigate rooms and perform elementary scientific tasks, demanding common-sense physical reasoning across 7 topics. Results are in \cref{tab:wm_science_world}.
\begin{table}[ht]
  \centering
  \resizebox{\textwidth}{!}{%
  \begin{tabular}{lccccccc}
    \toprule
    \cmidrule(r){1-2}
    World Model & Search & Layout & Detail & Decision & Page Change & Other & Avg.\\
    \midrule
    \rowcolor{gray!15}
    \multicolumn{8}{c}{\bf Baseline Models} \\
    \midrule
    Llama3.1-8B & 0.0 & 2.9 & 10.2 & 50.0 & 2.7 & 29.1 & 28.9\\
    Qwen3-8B & 0.0 & 0.0 & 14.3 & 39.5 & 2.7 & 32.6 & 25.1\\
    Qwen3-14B & 0.0 & 2.9 & 12.2 & 46.2 & 2.7 & 31.9 & 27.9\\
    Phi-4-mini & 0.0 & 0.0 & 4.1 & 50.0 & 2.7 & 11.3 & 24.9\\
    GPT-oss-20B & 0.0 & 0.0 & 16.3 & 50.0 & 5.4 & 0.7 & 24.0\\
    GPT-5-mini & 89.4 & 94.1 & 81.6 & 66.9 & 94.6 & 96.5 & 81.4\\
    \midrule
    \rowcolor{gray!15}
    \multicolumn{8}{c}{\bf Llama3.2-1B} \\
    \midrule
    SFT (100\% data) & 0.0 & 58.8 & 20.4 & 50.0 & 10.7 & \textbf{100.0} & 47.5\\
    \midrule
    Neural WM (Phase 1) & 0.0 & 0.0 & 0.0 & 50.0 & 0.9 & 0.7 & 22.2\\
    Symbolic WM (Phase 1) & \textbf{100.0} & 23.5 & 81.6 & \textbf{83.8} & 75.0 & 96.5 & 83.4\\
    WALL-E & 0.0 & 0.0 & 2.0 & 49.0 & 0.9 & 1.4 & 22.1 \\
    \ours\ (Phase 1) & 98.5 & 2.9 & 73.5 & \textbf{83.8} & 71.4 & 95.0 & 80.9\\
    \midrule
    Neural WM (Phase 2) & 0.0 & 52.9 & 19.2 & 50.0 & 12.4 & \textbf{100.0} & 45.9\\
    Symbolic WM (Phase 2) & \textbf{100.0} & 88.2 & \textbf{98.0} & \textbf{83.8} & \textbf{100.0} & 96.5 & 91.5\\
    WALL-E & 0.0 & 38.2 & 18.4 & 50.0 & 8.9 & \textbf{100.0} & 46.1 \\
    \ours~(Phase 2, 60\% data) & \textbf{100.0} & \textbf{91.2} & 95.9 & \textbf{83.8} & \textbf{100.0} & \textbf{100.0} & \textbf{92.2} \\
    \midrule
    \rowcolor{gray!15}
    \multicolumn{8}{c}{\bf Qwen3-4B} \\
    \midrule
    SFT (100\% data) & 0.0 & 64.7 & 18.4 & 52.5 & 11.6 & 92.2 & 47.3 \\
    \midrule
    Neural WM (Phase 1) & 0.0 & 0.0 & 16.3 & 50.0 & 3.6 & 27.0 & 28.9 \\
    Symbolic WM (Phase 1) & \textbf{100.0} & 23.5 & 73.5 & 66.6 & 33.0 & 97.2 & 68.9\\
    WALL-E & 0.0 & 0.0 & 14.3 & 50.0 & 2.7 & 35.5 & 30.3 \\
    \ours\ (Phase 1) & 97.0 & 2.9 & 65.3 & 83.8 & 29.5 & 99.3 & 74.4 \\
    \midrule
    Neural WM (Phase 2) & 0.0 & 64.7 & 19.2 & 51.0 & 13.5 & 93.1 & 46.1 \\
    Symbolic WM (Phase 2) & \textbf{100.0} & \textbf{100.0} & \textbf{100.0} & \textbf{83.8} & 95.5 & 92.9 & 90.8 \\
    WALL-E & 0.0 & 58.8 & 14.3 & 50.0 & 9.8 & 96.5 & 46.2 \\
    \ours~(Phase 2, 60\% data) & \textbf{100.0} & \textbf{100.0} & 95.9 & \textbf{83.8} & \textbf{100.0} & \textbf{100.0} & \textbf{92.6} \\
    \bottomrule
  \end{tabular}
  }
  \caption{
  World modeling accuracy on WebShop. \textbf{Bold} numbers highlight the best performance for each backbone model; data percentages give the fraction of the full training split used for neural fine-tuning.
  }
  \label{tab:wm_webshop}
\end{table}
\paragraph{\ours outperforms full fine-tuning with fewer training samples.} For Llama-1B, \ours achieves 68.3\% using only 45\% of the training data, surpassing SFT (64.4\%) on the full dataset. Similarly, for Qwen-4B, \ours (71.0\%) outperforms SFT (68.3\%), validating that a significant portion of trajectory data is redundant once the underlying deterministic rules are extracted. In other words, a large part of training data can be replaced by symbolic rules.

\paragraph{Reciprocal refinement improves both WMs, and their combination outperforms each in isolation.} Both WMs improve significantly from Phase~1 to Phase~2. Notably, as Neural WM improves, it exposes subtler failure modes that Symbolic WM captures in the second iteration, preventing the modules from becoming redundant.\footnote{We find marginal improvement when iterating the rule set over the same error clusters repeatedly.} After refinement, the combined \ours still surpasses each individual WM, confirming their complementary strengths: Symbolic WM scores deterministic transitions while Neural WM captures semantic information.

\paragraph{Context-based rule injection is insufficient for smaller models.} The WALL-E baseline, which injects symbolic rules into the LLM's input context as natural-language instructions, yields only marginal gains over the standalone Neural WM in Phase~1 (Llama: 38.3\% vs.\ 32.3\%; Qwen: 50.8\% vs.\ 46.9\%), while \ours achieves substantially larger improvements (48.8\% and 58.6\%). After fine-tuning in Phase~2, WALL-E improves considerably (Llama: 59.5\%; Qwen: 68.1\%) yet still trails \ours by 8.8 and 2.9 percentage points, respectively. The performance gap is notably wider for the smaller Llama-1B backbone, consistent with our hypothesis that smaller or heavily fine-tuned models tend to ignore prompt-injected rules. In contrast, \ours applies rule scores directly, making rule application independent of the model's instruction-following quality.

\paragraph{Scope of the GPT-5-mini comparison.} Phase~1 can surpass GPT-5-mini without neural training (e.g., both backbones on PlanCraft), but does not do so universally. After targeted training, Phase~2 \ours exceeds GPT-5-mini in all six backbone--environment settings. This comparison is conservative at deployment: GPT-5-mini answers each test input directly, whereas \ours uses it only during offline rule induction; the resulting rules are then frozen. Weak and same-backbone generator controls in \cref{app:new-results} further separate score-level integration from teacher-model knowledge transfer.

\begin{table}[ht]
  \centering
  \resizebox{\textwidth}{!}{%
  \begin{tabular}{lccccc}
    \toprule
    \cmidrule(r){1-2}
    World Model & Smelt & Move-Easy & Move-Medium & Move-Hard & Average\\
    \midrule
    \rowcolor{gray!15}
    \multicolumn{6}{c}{\bf Baseline Models} \\
    \midrule
    Llama3.1-8B & 90.6 & 91.2 & 69.5 & 36.5 & 72.6\\
    Qwen3-8B & 48.4 & 90.0 & 68.4 & 33.9 & 68.0\\
    Qwen3-14B & 56.3 & 91.5 & 64.4 & 35.7 & 68.9\\
    Phi-4-mini & 75.0 & 90.5 & 72.4 & 33.0 & 70.8\\
    GPT-oss-20B &62.5 & 81.3 & 70.7 & 27.4 & 63.7\\
    GPT-5-mini & 93.8 & 91.5 & 60.3 & 47.0 & 73.8 \\
    \midrule
    \rowcolor{gray!15}
    \multicolumn{6}{c}{\bf Llama3.2-1B} \\
    \midrule
    SFT (100\% data) & 31.3 & \textbf{98.8} & 73.6 & 67.0 & 80.5 \\
    \midrule
    Neural WM (Phase 1) & 54.7 & 88.1 & 69.5 & 28.7 & 66.4 \\
    Symbolic WM (Phase 1) & 45.3 & 93.4 & 47.7 & 48.7 & 69.2 \\
    WALL-E & 54.7 & 87.8 & 66.1 & 29.1 & 65.8 \\
    \ours\ (Phase 1) & 71.9 & 83.9 & 92.5 & \textbf{69.6} & 81.0 \\
    \midrule
    Neural WM (Phase 2) & 34.4 & 97.8 & 75.3 & 47.0 & 75.4 \\
    Symbolic WM (Phase 2) & 81.3 & 89.1 & 87.9 & 62.2 & 81.2 \\
    WALL-E & 35.9 & 97.8 & 76.4 & 43.9 & 75.0 \\
    \ours~(Phase 2, 35\% data) & \textbf{98.4} & 95.1 & \textbf{93.7} & 67.0 & \textbf{87.7} \\
    \midrule
    \rowcolor{gray!15}
    \multicolumn{6}{c}{\bf Qwen3-4B} \\
    \midrule
    SFT (100\% data) & 87.5 & 99.8 & 94.8 & \textbf{70.0} & \textbf{90.1} \\
    \midrule
    Neural WM (Phase 1) & 57.8 & 89.3 & 69.5 & 31.3 & 67.9 \\
    Symbolic WM (Phase 1) & 85.9 & 79.3 & 82.2 & 66.1 & 76.9 \\
    WALL-E & 40.6 & 88.8 & 63.2 & 33.9 & 65.9 \\
    \ours\ (Phase 1) & \textbf{93.8} & 83.0 & 92.5 & 67.4 & 81.6 \\
    \midrule
    Neural WM (Phase 2) & 87.5 & \textbf{100.0} & 93.1 & 51.7 & 85.1 \\
    Symbolic WM (Phase 2) & 81.3 & 89.8 & 86.8 & 45.7 & 77.0 \\
    WALL-E & 90.6 & 99.5 & 87.9 & 53.5 & 84.5 \\
    \ours~(Phase 2, 35\% data) & 92.2 & 97.1 & \textbf{96.6} & 65.6 & 88.4 \\
    \bottomrule
  \end{tabular}
  }
  \caption{
  World modeling accuracy on PlanCraft. \textbf{Bold} numbers highlight the best performance for each backbone model; data percentages give the fraction of the full training split used for neural fine-tuning.
  }
  \label{tab:wm_plan_craft}
\end{table}
\subsection{WebShop}\label{sec:webshop}
WebShop is an e-commerce environment that requires the agent to navigate webpages, search and purchase items that satisfy particular requirements, demanding both rigid web navigation protocols and semantic understanding of product details. We categorize tasks into six action types (see \cref{app:webshop-types}). \cref{tab:wm_webshop} summarizes the results.

\paragraph{Symbolic rules can vastly outperform LLMs on structural tasks.} The ``Search'' and ``Decision'' tasks require exact logic verification. Aside from \texttt{gpt-5-mini}, all neural models score zero on ``Search'' and perform random guessing on ``Decision'', whereas Symbolic WM achieves 100\% and $>$66\% respectively even in Phase~1. This demonstrates that simple Python rules can solve deterministic constraints that completely baffle fine-tuned LLMs.

\paragraph{Synergy benefits rule-solvable tasks and tolerates weak Neural WMs.} While ``Layout'' can be fully resolved by rules (Phase~2 Symbolic WM achieves 100\% for Qwen), it is difficult to induce such rules from the initialized model's development-set errors (Phase~1: 23.5\%). By learning from the mistakes of a fine-tuned Neural WM, Phase~2 Symbolic WM improves significantly, confirming that each module learns better from the other's residual errors. We also note that in the Llama Phase~1 results, \ours (80.9\%) slightly underperforms Symbolic WM alone (83.4\%) due to the extremely weak initialized Neural WM (22.2\%) acting as noise. However, once the Neural WM reaches moderate competency after fine-tuning (Phase~2: 45.9\%), the combined framework (92.2\%) overtakes Symbolic WM (91.5\%).

\subsection{PlanCraft}\label{sec:Plancraft}
PlanCraft is a Minecraft-based crafting environment, which requires world models to possess game-specific knowledge, i.e., the crafting recipes (or infer them from common sense), and strictly adhere to the rules of the game. Tasks are categorized into \textbf{Smelt} (furnace interactions) and \textbf{Move} (inventory manipulation), with Move further split into Easy, Medium, and Hard (see \cref{app:plancraft-types}). \cref{tab:wm_plan_craft} presents the results.

\paragraph{Fine-tuning tradeoffs highlight the value of hybrid modeling.} For Llama, SFT drops \textit{Smelt} accuracy from 54.7\% to 31.3\%, indicating training instability and possible forgetting of furnace logic. In contrast, \ours achieves 98.4\% on Smelt (Llama) and 92.2\% (Qwen); explicit rules preserve relevant transition logic even when the neural component degrades.

\paragraph{Medium-level tasks benefit most from neuro-symbolic synergy.} ``Move-Medium'' tasks require both recipe knowledge (symbolic) and goal-oriented planning (neural). \ours achieves 96.6\% for Qwen, outperforming both Neural WM (93.1\%) and Symbolic WM (86.8\%) in isolation. ``Move-Easy'' tasks are trivial for the LLM alone (98.8\% SFT), while ``Move-Hard'' tasks involve complex dynamics that resist simple rule encoding, suggesting our framework delivers the highest value at the intersection of planning and rigid logic.

\paragraph{Low-data evaluation.} With only 100 PlanCraft pairs randomly drawn from the stricter rule-uncovered training pool, Llama-1B \ours reaches 86.1\% test accuracy, close to 87.7\% from the full pipeline using 35\% of the 6,083 available pairs. Only the neural fine-tuning subset is restricted to 100 pairs; rule induction uses the same development split as the full-data experiment. This result establishes a concrete low-data point on PlanCraft, rather than a general few-shot claim across all environments.

\paragraph{Open-ended WebShop.} We further test whether the learned WM helps when next-state choices are not predefined. For each retained clickable action, a Llama-1B WM samples one successor and scores it using token log-likelihood and, when enabled, executable rules. The three highest-scoring action--successor pairs advise a WebShop-7B-SFT agent~\citep{xia2026skillrl}, which makes the final choice rather than executing the WM argmax. \Cref{tab:open-ended} shows that rule-guided lookahead improves average reward from 0.2755 to 0.3276 over 100 sessions. A same-protocol no-rule lookahead control did not improve over the base agent. Full generation and filtering settings are in \cref{app:new-results}.

\begin{table}[ht]
  \centering
  \begin{tabular}{lc}
    \toprule
    Method & Average reward \\
    \midrule
    Base WebShop-7B-SFT agent & 0.2755 \\
    Rule-guided one-step lookahead & \textbf{0.3276} \\
    \bottomrule
  \end{tabular}
  \caption{Open-ended WebShop evaluation. Both rows use 100 sessions with a 15-step cap.}
  \label{tab:open-ended}
\end{table}

\subsection{Robustness controls}\label{sec:robustness}
\Cref{tab:main-controls} summarizes key ablations and controls; complete per-environment results and definitions are in \cref{app:new-results}.

\begin{table}[ht]
  \centering
  \begin{tabular}{lcc}
    \toprule
    Variant & Llama3.2-1B & Qwen3-4B \\
    \midrule
    Full SFT & 64.1 & 68.6 \\
    \ours{} (adaptive, released rules) & \textbf{82.7} & 83.7 \\
    Linear interpolation & 82.4 & \textbf{84.5} \\
    Fixed scale $\gamma=1$ & 70.1 & 73.9 \\
    GPT-5-nano rule generator & 78.8 & 79.3 \\
    WALL-E-2.0-style feedback & 49.0 & 56.5 \\
    \bottomrule
  \end{tabular}
  \caption{Additional controls: mean test accuracy (\%) across the three environments. Score-integration rows use the released final rule snapshots. WALL-E-2.0-style is an executable-feedback implementation, not an exact reproduction.}
  \label{tab:main-controls}
\end{table}

\paragraph{Interpretation.} Linear interpolation closely matches log-linear reranking (82.4/84.5 versus 82.7/83.7 average accuracy), confirming that the algebraic combiner is not the source of novelty; fixed $\gamma=1$ is markedly weaker (70.1/73.9), showing the importance of candidate-adaptive scaling. Rules generated by GPT-5-nano remain above full SFT in most settings, and using Qwen3-4B to generate rules for its own WM still reaches 87.1\% on PlanCraft and 80.3\% on WebShop, reducing the teacher-knowledge confound. In contrast, the WALL-E-2.0-style executable-feedback baseline averages only 49.0/56.5, showing that direct score integration outperforms this context-feedback implementation in our 1B/4B setting. Rule-set size, selection protocol, cost, and per-environment results are detailed in \cref{app:new-results}.

\section{Conclusion and limitations} \label{sec:conclusion}
We introduced \ours, a framework that combines the semantic flexibility of LLM world models with executable rules through candidate-level log-linear reranking. Rule scores impose strong soft penalties without relying on instruction following, while reciprocal refinement focuses neural training and rule induction on complementary cases. Results across three environments and an open-ended WebShop study support its effectiveness and data efficiency. Limitations include environment-specific rules, sensitivity to backbone and rule-generator capability, rule selection on the development set without a second held-out validation split, and open-ended evaluation limited to one-step lookahead on one environment. Longer-horizon closed-loop planning and separately held-out rule selection are important future work.

\bibliography{wm}
\bibliographystyle{colm2026_conference}

\newpage
\appendix
\section{Additional results and analysis}\label{app:new-results}

\subsection{Low-data and open-ended evaluations}
The full training-set sizes are reported in the main text. To test a small neural-training regime, we randomly sampled 100 $(s,a,s')$ pairs from the 1,349 candidate PlanCraft transition-training rows remaining after the train split and strict removal of rule-covered pairs (seed 43) for Phase~2 Llama3.2-1B fine-tuning; rule induction used the same development split as in the full-data experiment. This stricter pool is distinct from the full pipeline's 35\% subset, which also resamples covered rows. The 100-pair run achieves 86.1\% test accuracy, close to the full pipeline's 87.7\%. This result is limited to PlanCraft and should not be read as a three-environment few-shot study.

We also evaluate whether the world model can aid an agent without predefined next-state choices. We use the WebShop-7B-SFT agent from SkillRL~\citep{xia2026skillrl} and one-step lookahead. In preliminary trials, simply inserting a predicted state into the agent context was unreliable; instead, we use the (possibly rule-modified) likelihood of that prediction as an action-quality signal. \Cref{tab:open-ended} shows that rule-guided lookahead improves average reward over the base agent, whereas a same-protocol no-rule control does not.

The base and rule-guided runs evaluate WebShop sessions 0--99 with a 15-step cap and seed 0. For rule-guided lookahead, a Llama3.2-1B WM fine-tuned on the rule-guided subset samples one successor for each retained clickable action (temperature 0.8, top-$p$ 0.95, and at most 10 actions). Before successor sampling, rules remove an action if their aggregate score for the hypothetical terminal candidate \texttt{Fail} is positive. We rank the remaining actions by the unnormalized token log-likelihood of the sampled successor plus the sum of 20 unit-weight rule scores. The top three action--successor pairs and their scores are supplied as advisor context to the 7B agent, which makes the final choice (temperature 0.2); thus the controller does not directly execute the top-ranked action. Advice is recomputed after every observed environment transition, so episodes are closed-loop although the WM lookahead horizon is one step. The no-rule ablation disables the rule filter and additive rule scores.

\subsection{Score integration and scaling}
We compare our log-linear score with three alternatives suggested during review. Linear interpolation uses
\begin{equation*}
p'_i=(1-\alpha)p_i+\alpha\frac{\exp(\gamma E_i)}{\sum_k\exp(\gamma E_k)}.
\end{equation*}
Hard masking retains $M=\{i:\gamma E_i\geq\max_k\gamma E_k-m\}$ and assigns $\log p_i$ to candidates in $M$ and $-\infty$ otherwise. For product-of-experts, $q_{ij}=\operatorname{clip}((e_{ij}+1)/2,\epsilon,1-\epsilon)$ and $\log p'_i=\log p_i+\gamma\sum_jw_j\log q_{ij}$. For the alternative combiners, combiner-specific hyperparameters, including fixed versus candidate-adaptive scaling, are selected on the development set; the log-linear row uses the candidate-adaptive scale defined in the main method.

\begin{table}[H]
  \centering
  \resizebox{\textwidth}{!}{%
  \begin{tabular}{llcccc}
    \toprule
    Backbone & Combiner & PlanCraft & ScienceWorld & WebShop & Avg. \\
    \midrule
    Llama3.2-1B & Log-linear & \textbf{87.7} & \textbf{68.3} & \textbf{92.2} & \textbf{82.7} \\
    & Product-of-experts & 84.5 & 67.5 & \textbf{92.2} & 81.4 \\
    & Hard mask & 86.9 & 67.9 & \textbf{92.2} & 82.3 \\
    & Linear interpolation & 87.4 & 67.7 & \textbf{92.2} & 82.4 \\
    \midrule
    Qwen3-4B & Log-linear & 87.9 & 70.6 & 92.6 & 83.7 \\
    & Product-of-experts & 87.4 & \textbf{71.2} & 92.0 & 83.5 \\
    & Hard mask & 87.1 & 71.0 & \textbf{92.7} & 83.6 \\
    & Linear interpolation & \textbf{89.8} & 71.0 & \textbf{92.7} & \textbf{84.5} \\
    \bottomrule
  \end{tabular}
  }
  \caption{Test accuracy for candidate-score combination functions using the released final rule snapshots. Bold marks the best value within each backbone and column (ties included). Log-linear and linear interpolation are comparable, confirming that the algebraic combiner is not the source of novelty. The Qwen log-linear values differ slightly from the independently produced main-run snapshots.}
  \label{tab:fusion}
\end{table}

Multiplying a fixed $\gamma$ by a constant can be absorbed into the learned weights. The meaningful design choice here is instead whether the scale adapts to each candidate set. \Cref{tab:gamma} compares $\gamma=1$ with $\gamma=1+\max_k\log p_k-\min_k\log p_k$. Adaptive scaling is especially important when the neural scores are misleading on WebShop.

\begin{table}[H]
  \centering
  \resizebox{\textwidth}{!}{%
  \begin{tabular}{llcccc}
    \toprule
    Backbone & Scale & PlanCraft & ScienceWorld & WebShop & Avg. \\
    \midrule
    Llama3.2-1B & $\gamma=1$ & 83.4 & 65.6 & 61.3 & 70.1 \\
    & $\gamma=\text{gap}+1$ & \textbf{87.7} & \textbf{68.3} & \textbf{92.2} & \textbf{82.7} \\
    \midrule
    Qwen3-4B & $\gamma=1$ & 87.4 & \textbf{71.6} & 62.7 & 73.9 \\
    & $\gamma=\text{gap}+1$ & \textbf{87.9} & 70.6 & \textbf{92.6} & \textbf{83.7} \\
    \bottomrule
  \end{tabular}
  }
  \caption{Sensitivity to the rule-score scaling design on the released final rule snapshots. ``gap'' is the largest log-likelihood gap in the candidate set.}
  \label{tab:gamma}
\end{table}

\subsection{Rule-generator controls}
\Cref{tab:weak-generator} replaces GPT-5-mini with the weaker GPT-5-nano rule generator. Performance decreases, but remains above the base model in every setting and above full SFT in four of six settings. Thus, the framework is not solely dependent on GPT-5-mini, although generator quality matters, particularly on WebShop.

\begin{table}[H]
  \centering
  \resizebox{\textwidth}{!}{%
  \begin{tabular}{llcccc}
    \toprule
    Backbone & Method & PlanCraft & ScienceWorld & WebShop & Avg. \\
    \midrule
    Llama3.2-1B & Base model & 66.4 & 32.3 & 22.2 & 40.3 \\
    & Full SFT & 80.5 & 64.4 & 47.5 & 64.1 \\
    & \ours{} + GPT-5-mini rules & \textbf{87.7} & \textbf{68.3} & \textbf{92.2} & \textbf{82.7} \\
    & \ours{} + GPT-5-nano rules & 82.1 & 63.4 & 90.9 & 78.8 \\
    \midrule
    Qwen3-4B & Base model & 67.9 & 46.9 & 28.9 & 47.9 \\
    & Full SFT & \textbf{90.1} & 68.3 & 47.3 & 68.6 \\
    & \ours{} + GPT-5-mini rules & 87.9 & \textbf{70.6} & \textbf{92.6} & \textbf{83.7} \\
    & \ours{} + GPT-5-nano rules & 83.7 & 69.7 & 84.5 & 79.3 \\
    \bottomrule
  \end{tabular}
  }
  \caption{Test accuracy with a weaker rule generator. The reported GPT-5-nano results use the same adaptive scaling as the main method.}
  \label{tab:weak-generator}
\end{table}

For the cleanest knowledge-transfer control, we also use Qwen3-4B itself to generate rules for the Qwen3-4B world model. \Cref{tab:self-generator} shows that same-backbone generation remains effective on both tested environments. The larger drop and lower accepted-rule count on WebShop also expose a real dependence on generator capability; this control was not run on ScienceWorld or with Llama3.2-1B.

\begin{table}[H]
  \centering
  \begin{tabular}{llcc}
    \toprule
    Environment & Rule generator & Accepted rules & Test accuracy \\
    \midrule
    PlanCraft & GPT-5-mini & 30 & 87.4 \\
    & GPT-5-nano & 24 & 87.3 \\
    & Qwen3-4B (same backbone) & 20 & 87.1 \\
    \midrule
    WebShop & GPT-5-mini & 30 & 92.6 \\
    & GPT-5-nano & 14 & 84.5 \\
    & Qwen3-4B (same backbone) & 6 & 80.3 \\
    \bottomrule
  \end{tabular}
  \caption{Same-backbone rule-generator controls for the Qwen3-4B neural WM. PlanCraft uses fixed $\gamma=1$ for all three rows; WebShop uses adaptive scaling.}
  \label{tab:self-generator}
\end{table}

\subsection{Executable-feedback baseline}
WALL-E~2.0 uses executable functions but presents their outputs to an LLM through its context. We implement a \emph{WALL-E-2.0-style} baseline rather than claiming an exact reproduction: it executes the same learned Python rules on every candidate, converts their outputs into concise textual feedback, and injects up to five feedback items with the largest absolute rule scores per candidate. \Cref{tab:walle2} shows that this remains substantially weaker than direct score integration for our 1B/4B world models. One possible explanation for its regression relative to WALL-E~1.0 is that the executable-feedback prompt is more complex and the original WALL-E evaluations use much larger GPT-4 models with stronger instruction following.

\begin{table}[H]
  \centering
  \begin{tabular}{llcccc}
    \toprule
    Backbone & Method & PlanCraft & ScienceWorld & WebShop & Avg. \\
    \midrule
    Llama3.2-1B & WALL-E~1.0 & 75.0 & 59.5 & 46.1 & 60.2 \\
    & WALL-E-2.0-style & 56.3 & 39.3 & 51.4 & 49.0 \\
    & \ours & \textbf{87.7} & \textbf{68.3} & \textbf{92.2} & \textbf{82.7} \\
    \midrule
    Qwen3-4B & WALL-E~1.0 & 84.5 & 68.1 & 46.2 & 66.3 \\
    & WALL-E-2.0-style & 66.8 & 52.5 & 50.3 & 56.5 \\
    & \ours & \textbf{87.9} & \textbf{70.6} & \textbf{92.6} & \textbf{83.7} \\
    \bottomrule
  \end{tabular}
  \caption{Aggregate test accuracy for context-based rule injection and direct score integration.}
  \label{tab:walle2}
\end{table}

\subsection{Rule-set scale, evaluation protocol, and cost}
The six final GPT-5-mini rule files contain 216 rules (36 per model--environment setting on average) and 729,103 bytes of Python code (3,375 bytes per rule on average), adding no neural parameters. \Cref{tab:rule-stats} gives the per-setting counts and footprint. Rules are intentionally environment-specific but can transfer across tasks within an environment; for example, a ScienceWorld door-opening rule applies across task goals. Opposing rules can coexist because their scalar contributions partially cancel, while development-set filtering rejects rules that harm aggregate accuracy.

\begin{table}[H]
  \centering
  \begin{tabular}{llrr}
    \toprule
    Environment & Backbone & Rules & Code bytes \\
    \midrule
    ScienceWorld & Llama3.2-1B & 48 & 158,012 \\
    ScienceWorld & Qwen3-4B & 44 & 163,435 \\
    WebShop & Llama3.2-1B & 20 & 49,730 \\
    WebShop & Qwen3-4B & 30 & 54,934 \\
    PlanCraft & Llama3.2-1B & 44 & 185,988 \\
    PlanCraft & Qwen3-4B & 30 & 117,004 \\
    \midrule
    \multicolumn{2}{l}{Total} & 216 & 729,103 \\
    \bottomrule
  \end{tabular}
  \caption{Size of the final GPT-5-mini-generated symbolic WMs.}
  \label{tab:rule-stats}
\end{table}

To characterize rule behavior, \cref{tab:rule-diagnostics} reports descriptive diagnostics of the final, post-refinement rule sets on the development split. For an item whose gold next state is $y^*$, $k=\lvert\{j:f_j(b,a,y^*)\neq0\}\rvert$ and coverage is the fraction with $k>0$. For each rule--item pair that is active for at least one candidate, we define $\Delta=f_j(b,a,y^*)-\max_{y\neq y^*}f_j(b,a,y)$. Helpful, harmful, and tied events have positive, negative, and zero $\Delta$, respectively. PlanCraft's 286 single-candidate items contribute to coverage but not to the quality distribution.

\begin{table}[H]
  \centering
  \begingroup
  \small
  \setlength{\tabcolsep}{4pt}
  \begin{tabular}{@{}lrrrrrr@{}}
    \toprule
    Setting & Gold coverage & Mean $k$ & Helpful & Harmful & Tie & Rules H/Harm/T \\
    \midrule
    PC / Llama-1B & 876/876 & 33.45 & 25.98 & 14.03 & 59.99 & 38/6/0 \\
    PC / Qwen-4B & 876/876 & 23.65 & 22.46 & 14.35 & 63.19 & 24/6/0 \\
    SW / Llama-1B & 457/497 & 4.96 & 21.00 & 7.70 & 71.30 & 43/4/1 \\
    SW / Qwen-4B & 456/497 & 3.39 & 24.58 & 6.84 & 68.58 & 40/3/1 \\
    WS / Llama-1B & 636/638 & 2.94 & 75.13 & 8.84 & 16.03 & 20/0/0 \\
    WS / Qwen-4B & 638/638 & 9.06 & 49.81 & 8.11 & 42.08 & 30/0/0 \\
    \bottomrule
  \end{tabular}
  \endgroup
  \caption{Development-set diagnostics for each final rule set (PC: PlanCraft; SW: ScienceWorld; WS: WebShop). Helpful/harmful/tie columns are event percentages; the last column gives rule counts classified by mean $\Delta$. These diagnostics do not estimate the Phase~2 retained-data fraction, which used the initialized rules on training trajectories followed by stochastic $1/k$ resampling.}
  \label{tab:rule-diagnostics}
\end{table}

The development split is used for rule selection and coordinate-descent weight calibration. The final rules and weights are then frozen, and all reported test-performance numbers are computed on the unseen test split. Across the six final settings, learned weights improve development accuracy by 0.3--4.8 percentage points (2.6 on average) over uniform weights. Because repeated selection on one development set can still overfit, a second held-out rule-selection split is a useful future robustness check.

For a single PlanCraft run with Qwen3-4B, rule induction made 150 GPT-5-mini calls with about 2,700 tokens per call and cost less than USD~1 at the experiment-time API price. On four 48GB A6000 GPUs, full SFT took 2.3 wall-clock hours, whereas the reduced-data neural training stage of \ours took 0.8 hours. API cost and GPU time measure different resources and are therefore reported separately.

\subsection{Representative learned rule}\label{app:rule-example}
The following abridged PlanCraft rule illustrates the executable rules used by the symbolic WM; complete rules are in the released code.
\begin{verbatim}
def rule_reward(state, action, choice):
    src, dst, q = parse_move_action(action)
    if src is None:
        return 0.0                 # inactive for non-move actions
    old, new = parse_items(state), parse_items(choice)
    if src not in old or dst not in new:
        return -1.0                # confident legality violation
    if not valid_quantity_transfer(old, new, src, dst, q):
        return -1.0
    others_ok = unchanged_except(old, new, {src, dst, "[0]"})
    output_changed = old.get("[0]") != new.get("[0]")
    if not others_ok:
        return 0.3 if output_changed else -0.6
    return 1.0 if output_changed else 0.5
\end{verbatim}
It activates only for a matching action form, uses $-1$ for clear violations, and returns smaller-magnitude scores under uncertainty. A single induced rule need not encode a complete human-correct theory---for example, this rule mildly rewards an output-slot change---because other rules penalize changes to unrelated products and the weighted ensemble supplies the final score. The rules are environment-specific rather than universal. In the WebShop teaser, ``prev/next'' appears commonsensical, but the recorded next state after the feature-button click is ``back to Search/prev''; an exact environment rule distinguishes these otherwise plausible candidates.

\section{Data selection and routing analysis}\label{app:ablation}
\begin{figure}[ht]
\centering
\includegraphics[width=0.7\textwidth]{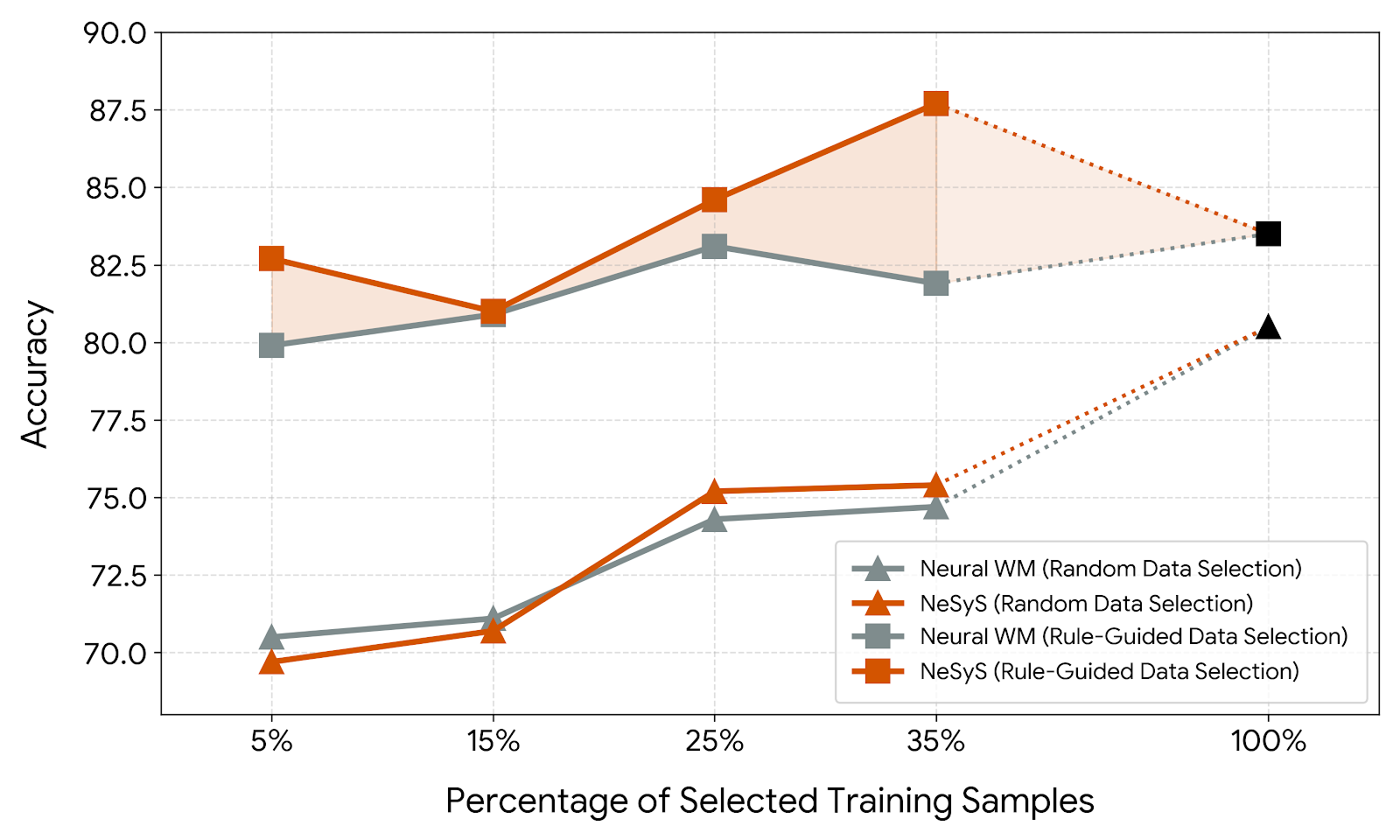}
\caption{Comparing rule-guided data selection (ours) vs. random data selection under different training budgets on PlanCraft. The backbone model is Llama-3.2-1B-Instruct. The shadow in the figure highlights the performance gain of our strategy.}
\label{fig:random_vs_rule_filtering}
\end{figure}

\paragraph{Ablation studies on data selection methods.} We compare the rule-guided data selection method we used in the main experiments against a random selection baseline. \cref{fig:random_vs_rule_filtering} shows that Neural WMs trained with two data selection methods perform similar without symbolic rules, but rule-guided method consistently perform better when paired with Symbolic WM. Additionally, we observe that the performance of Neural WM keeps improving until using all the training samples, but this is not true for \ours. These two observations further prove the complementary strengths of Neural and Symbolic WM in our framework, and the effectiveness of the proposed rule-guided data selection mechanism. 

\paragraph{Exploration of training a lightweight router.} 
Building on the neuro-symbolic synergy observed in our main experiments, we explored the potential of a routing mechanism to dynamically select the optimal configuration (i.e., Neural vs. Symbolic WM, phase 1 vs. phase 2 models, and the scaling factor $\gamma$). We compared a lightweight XGBoost router, trained on the development set, against an `oracle' router that acts as a theoretical upper bound by always selecting the optimal parameters. As detailed in \cref{app:router}, the XGBoost router yielded marginal gains over the baseline \ours. However, the oracle router achieved near-perfect scores, indicating that while simple routing is insufficient, developing a more sophisticated selection mechanism remains a high-potential direction for future work.

\section{Evaluation prompt}\label{app:prompt_eval}
When evaluating on ScienceWorld, we include recent trajectories and a memory of each visited location before the beginning of the trajectories, and ask for predicting the next observation, reward and difference of inventory. Here is the prompt we used:
\begin{tcolorbox}[
  title=Prompt used for Evaluating on ScienceWorld,
  colback=gray!5,
  colframe=black!50,
  fonttitle=\bfseries
]
\textbf{System:} You are a ScienceWorld transition model. Given the compressed history and ONE action, output exactly three lines in this order:\\
predicted\_observation: \textless text\textgreater\\
predicted\_reward: \textless float in [-1,1]\textgreater\\
predicted\_inventory\_diff: \textless one or more lines with '+ ' or '- ' prefixes only\textgreater
Do not output any additional text. \\[0.5em]
\textbf{User:} Task Description: \{task description\}\\
Memory (last look-around per visited location): \{memory\}\\
action: \{$a_k$\}\\
observation: \{$s_{k+1}$\}\\
reward: \{$r_{k}$\}\\
\ldots \\
current\_step\_action: \{$a_t$\}\\
\end{tcolorbox}

Here is the prompt we used for evaluating on WebShop:
\begin{tcolorbox}[
  title=Prompt used for Evaluating on WebShop,
  colback=gray!5,
  colframe=black!50,
  fonttitle=\bfseries
]
\textbf{System:} You are a transition model for the WebShop text environment. \\
Given the current page text (state) and the action taken, predict the next page text. \\
Your answer must be ONLY the next state's text, with no extra commentary. \\[0.5em]
\textbf{User:} Current state (page text): \{state\} \\
Action taken: \{$a_t$\}\\
Question: What is the next state's page text? Answer with ONLY the next state's text. 
\end{tcolorbox}

Here is the prompt we used for evaluating on PlanCraft:
\begin{tcolorbox}[
  title=Prompt used for Evaluating on PlanCraft,
  colback=gray!5,
  colframe=black!50,
  fonttitle=\bfseries
]
\textbf{System:} You are a Minecraft transition model. Given a current state and ONE action, return the next state text in the exact format used by the data (same headers and lists). Do not output actions. Enforce environment rules: cannot move/smelt into [0]; crafting outputs appear at [0] and must be moved to an inventory slot to complete. Output only the next state text. \\[0.5em]
\textbf{User:} \{state\} \\
Action: \{$a_t$\}
\end{tcolorbox}

\section{Rule induction prompts}\label{app:prompt_rule_induction}
Here are the prompts we used for rule induction in each environment. Example programs are omitted for clarity; they are Python programs that implement the example rules and can be found in our published code.

\begin{tcolorbox}[
  title=Rule induction for WebShop,
  colback=gray!5,
  colframe=black!50,
  fonttitle=\bfseries,
  breakable
]
\textbf{User:}

Analyze the following WebShop transition model error cases and summarize one actionable improvement rule. Follow these guidelines:

\textbf{[Error Cases]} \\
\{cases\}

\textbf{[WebShop Format]}
\begin{itemize}
  \item The state text is the current WebShop page content.
  \item The action is an environment action such as:
  \begin{itemize}
    \item \texttt{click[buy now]}
    \item \texttt{click[}$<$\texttt{ prev]}
    \item \texttt{search[...]}
  \end{itemize}
  \item The choice is a candidate next state or a terminal token: Success or Fail.
\end{itemize}

\textbf{[Analysis Requirements]}
\begin{enumerate}
  \item Identify shared action types across cases.
  \item Infer what the correct next page or terminal result should be.
  \item Formulate one generalizable and checkable rule returning a score in $[-1,1]$.
  \item Apply the rule only when its conditions match; otherwise return 0.0.
  \item Phrase the rule after \#\#\# Rule \#\#\#.
  \item Write a Python program after \#\#\# Program \#\#\#.
\end{enumerate}

\textbf{[Example Rules and Programs]}

1) Example rule (Buy Now missing $\Rightarrow$ Fail): If action is exactly click[buy now] but the current page text does NOT contain a Buy Now button,
   then the correct terminal result is Fail. Return 1 if choice is Fail and -1 otherwise.

Example Program: (Omitted)

2) Example rule (\textless{} Prev $\Rightarrow$ back to results): If action is click[\textless{} prev] and the current page text contains a \textless{} Prev button,
   then the next page is usually a search results list page that contains 'Total results' and a 'Next \textgreater{}' button.

Example Program: (Omitted)
\end{tcolorbox}

\begin{tcolorbox}[
  title=Rule induction on PlanCraft,
  colback=gray!5,
  colframe=black!50,
  fonttitle=\bfseries,
  breakable
]
\textbf{User:}

Analyze the following model error cases and summarize one actionable improvement rule. Follow these guidelines:

\textbf{[Error Cases]} \\
\{cases\}

\textbf{[Analysis Requirements]}
\begin{enumerate}
  \item Try to find patterns in the questions. What do they have in common? Are the actions of the same type? Do the states share similarities?
  \item Try to find patterns in the correct answers. What are their shared characteristics?
  \item Try to find patterns in the incorrect answers. What makes them incorrect?
  \item Formulate one generalizable rule for the presented error cases. The rule should be detailed enough to be programmed and used to score each candidate choice. It should apply only when the shared patterns are observed, encouraging correct patterns and discouraging incorrect ones.
  \item If a detailed rule cannot be found for all error cases, describe a rule for a subset rather than being vague.
  \item Phrase the rule after \#\#\# Rule \#\#\#.
  \item Write a Python program after \#\#\# Program \#\#\#. The program should define a function rule\_reward(state, action, choice) that returns a float in $[-1,1]$, indicating how likely the choice is correct.
\end{enumerate}

\textbf{[Example Rule and Programs]}

\textbf{Example rule 1.} For a move action "move: from [I?] to [A/B/C?] with quantity q", the next state should reflect only the intended move. Penalize choices that change unrelated item counts.

\textbf{Example program:} (Omitted)

\textbf{Example rule 2.} For illegal actions other than move or smelt, the next state should not change.

\textbf{Example program:} (Omitted)

\end{tcolorbox}

\begin{tcolorbox}[
  title=Rule induction for ScienceWorld,
  colback=gray!5,
  colframe=black!50,
  fonttitle=\bfseries,
  breakable
]
\textbf{User:}

Analyze the following ScienceWorld model error cases and summarize one actionable improvement rule. Follow these guidelines:

\textbf{[Error Cases]} \\
\{cases\}

\textbf{[ScienceWorld Format]}
\begin{itemize}
  \item The state text is a compressed history including task description, optional memory, per-step history (action, observation, reward), and inventory or inventory\_diff snippets. It ends with current\_step\_action: \textless action\textgreater.
  \item The choice must contain exactly the following lines (order fixed):
  \begin{itemize}
    \item predicted\_observation: \textless text\textgreater
    \item predicted\_reward: \textless float in [-1,1]\textgreater
    \item predicted\_inventory\_diff: \textless zero or more +/- lines\textgreater
  \end{itemize}
\end{itemize}

\textbf{[Analysis Requirements]}
\begin{enumerate}
  \item Identify common action types (e.g., open X, pick up Y, use thermometer on Z).
  \item Infer what consistent predictions should look like for these actions.
  \item Design a rule that checks these consistencies by parsing the choice and, if needed, extracting the current action from the state.
  \item The rule should be specific and return a score in $[-1,1]$.
  \item Phrase the rule after \#\#\# Rule \#\#\#.
  \item Write a Python program after \#\#\# Program \#\#\# defining \texttt{rule\_reward(state, action, choice)}.
\end{enumerate}

\textbf{[Example Rules and Programs]}

1)Example rule (Pick up): If action starts with 'pick up \textless obj\textgreater', then:
   - predicted\_observation contains 'You move the \textless obj\textgreater{} to the inventory.'
   - predicted\_inventory\_diff contains a line starting with '+ ' and mentioning \textless obj\textgreater.
   
Example Program: (Omitted)

2) Example rule (Open): If action matches 'open \textless obj\textgreater', require:
   - predicted\_observation contains 'The \textless obj\textgreater{} is now open.' (or 'already open')
   - predicted\_inventory\_diff is empty or contains no changes

Example Program: (Omitted)

3) Example rule (Thermometer): If action contains 'use thermometer' and 'on \textless target\textgreater', then:
   - predicted\_observation contains 'the thermometer measures a temperature of'
   - predicted\_inventory\_diff should be empty

Example Program: (Omitted)
\end{tcolorbox}
We use the following prompt for reflection in rule induction:
\begin{tcolorbox}[
  title=Prompt for reflection in rule induction,
  colback=gray!5,
  colframe=black!50,
  fonttitle=\bfseries,
  breakable
]
\textbf{User:}

The following rule is causing negative impacts on some questions that were originally answered correctly. Please refine the rule to avoid these negative impacts while maintaining its beneficial effects.

\textbf{[Current Rule Description]} \\
\{rule\_description\}

\textbf{[Current Rule Program]}
\{current\_rule\}

\textbf{[Negative Impact Cases]}
\{negative\_impacted cases\}

\textbf{[Refinement Requirements]}
\begin{enumerate}
  \item Analyze why the current rule is causing originally correct answers to become wrong.
  \item Identify the specific conditions or patterns responsible for the negative impact.
  \item Refine the rule to be more precise and avoid these false positives.
  \item Maintain the beneficial effects the rule was designed to achieve.
  \item Make the refined rule more conservative to avoid breaking correct predictions.
  \item Write the refined rule after \#\#\# Rule \#\#\#.
  \item Write the refined Python program after \#\#\# Program \#\#\#. The program should define rule\_reward(state, action, choice) and return a float in $[-1,1]$.
\end{enumerate}

\textbf{[Focus Areas for Refinement]}
\begin{itemize}
  \item Add more specific conditions to prevent false positives.
  \item Consider edge cases that may be incorrectly handled.
  \item Make the rule more conservative in its judgments.
  \item Add additional validation checks before applying penalties or rewards.
\end{itemize}
\end{tcolorbox}

\section{Clustering details}\label{app:clustering}
We cluster the error cases based on the embeddings of the following part of the case respectively:
\begin{itemize}
    \item question
    \item correct answer
    \item wrongly selected answer
    \item action
    \item task name
\end{itemize}
To convert text to embedding, we first concatenate a tf-idf embedding of the selected text with the embedding generated by \texttt{all-MiniLM-L6-v2} from sentence transformers~\citep{reimers-2019-sentence-bert}. We reduce the dimension to 50 with a truncated SVD, then to 5 with a UMAP. We use the OPTICS algorithm with min samples=3, xi=0.05 and min cluster size=0.1 for clustering. 

\section{Router training}\label{app:router}
We train the XGBoost router with the following features:
\begin{itemize}
    \item Scores provided by each rule
    \item Logprob of each options provided by the base Neural WM
    \item Length of the question
    \item Length of the action
    \item task name
\end{itemize}
We try all combinations of the features and select the model with best dev set accuracy. We use a fixed set of hyperparameters: n estimators=2500, max depth=6, learning rate=0.01.

The detailed result is shown in \cref{tab:router}. The analysis can be found at \cref{app:ablation}.
\begin{table}[ht]
  \centering
  \begin{tabular}{lcccc}
    \toprule
    \cmidrule(r){1-2}
    Model & ScienceWorld & WebShop & PlanCraft & Average\\
    \midrule
    \multicolumn{5}{c}{\bf Llama3.2-1B} \\
    \midrule
    \ours & 68.3& 92.2& 87.7 & 82.7 \\
    XGBoost router & 68.7 & 92.4 &88.6 & 83.2\\    
    Oracle router & 82.9 & 94.3 & 95.8 & 91.0 \\
    \midrule
    \multicolumn{5}{c}{\bf Qwen3-4B} \\
    \midrule
    \ours & 71.0& 92.6& 88.4& 84.0 \\
    XGBoost router & 71.0 & 91.3&91.5 & 84.6\\   
    Oracle router & 84.2 & 99.9 & 98.2 & 94.1 \\
    \bottomrule
  \end{tabular}
  \caption{Performance of an oracle router and an actually trained XGBoost router.}
  \label{tab:router}
\end{table}
\section{WebShop task type definitions}\label{app:webshop-types}
We divide the WebShop tasks into 6 types by actions:
\begin{itemize}
    \item Search (search[xxx]), where the model needs to predict the possible search result.
    \item Layout (click[item - xxx]), where the model needs to predict the layout of the item page.
    \item Detail (click[description/features/reviews]), where the model needs to predict the detailed descriptions of the product or the detailed user feedback of the product. These buttons are standard for all product pages, and the result format is fixed.
    \item Decision (click[buy now]), where the model needs to decide whether we successfully bought the product that meets the description. All the questions in this category have only 2 choices: Success and Fail.
    \item Page Change (click[\textless{} prev/next \textgreater{}/back to search]), where the model needs to understand the meaning of these buttons and the format of the target pages.
    \item Other, corresponding to all other actions, mainly include choosing the size/specification of the products.
\end{itemize}

\section{PlanCraft difficulty level definitions}\label{app:plancraft-types}
We divide the PlanCraft move tasks into 3 difficulty levels:
\begin{itemize}
    \item Easy, including simple moves with no side effect.
    \item Medium, corresponding to moves that generate the product required to reach the goal (not necessarily the final product). This is easier than hard as the goal is part of the belief state.
    \item Hard, including moves that generate a side product.
\end{itemize}

\section{Benchmark creation details}\label{app:data-collection}
As discussed in \cref{sec:exp_setup}, we create the benchmarks based on the official expert trajectories provided by the environment. Here we enclose the details for each environment.

For ScienceWorld, we sample at most 50 dev/test variations from each pre-defined task (there are 30 predefined tasks in the environment), and at most 5 positions from each variation. We create one question at each position. 

For WebShop, we do a random 90:5:5 train:dev:test split on the provided trajectories. We use every time step in the dev/test trajectories to create a question. Since all the trajectories end up successfully buying the desired item, we randomly select middle positions of each trajectory and insert a ``click[buy now]'' action, which will result in a "Fail" state. Similar operation is performed on the training set.

For PlanCraft, we sample at most 6 steps from each provided dev/test trajectory. We enriched the training trajectory by 
\begin{itemize}
    \item Using side products in existing trajectory as goal and truncate the trajectory up to that position.
    \item Starting from the existing setups and generating new trajectories by looping through all possible recipes.
    \item Modifying the starting inventory of existing trajectories by removing a required ingredient or swapping it with a similar item.
\end{itemize}

After we sample the questions and the correct answers, we use \texttt{Llama-3.2-1b-instruct} to answer questions and use the wrong ones as the distractors.

\section{Rigorous definition of POMDP}\label{app:pomdp}
In \cref{sec:problem}, we defined the problem informally for readability. Here we provide a rigorous definition of the POMDP problem we study. It is usually written as $\mathcal{M} = (S, A, T, R, \Omega, O)$, where $S$ denotes the latent state space, $A$ the action space,
$T : S \times A \to \Delta(S)$ the transition kernel, $R : S \times A \to \mathbb{R}$ the reward function, $\Omega$ the observation space, and $O : S \times A \to \Delta(\Omega)$ the observation kernel. In this work, both actions $a_t \in A$ and observations $\omega_t \in \Omega$ are represented as natural-language strings.

A \emph{world model} is defined as a predictive model that estimates the next observation $\omega_t$ and reward $r_t$ conditioned on the current action $a_t$
and belief state $b_t$. Following prior LLM-based agent work, we approximate the belief state by a textual context constructed from the task description $g$ and a truncated history of recent observations, actions, and rewards that fit within the model’s context window.
\end{document}